\documentclass{article}
\usepackage{ijcai20}

\usepackage{times}
\usepackage{soul}
\usepackage{url}
\usepackage[hidelinks]{hyperref}
\usepackage[utf8]{inputenc}
\usepackage[small]{caption}
\usepackage{graphicx}
\usepackage{amsmath}
\usepackage{amsthm}
\usepackage{booktabs}
\usepackage{algorithm}
\usepackage{algorithmic}
\usepackage{amsfonts}
\usepackage{comment}

\usepackage{multirow}

\usepackage{relsize} 
\usepackage{xcolor}
\usepackage{xspace}
\usepackage{subfig}
\usepackage{tablefootnote}
\usepackage{longtable}
\usepackage{threeparttable}
\usepackage{lipsum,afterpage,refcount}
\newcommand{\setfootnotemark}{%
  \refstepcounter{footnote}%
  \footnotemark[\value{footnote}]}
\graphicspath{{figures/}}

\newcommand{\frameworkName}{\textsc{DArtNet}\xspace}
\newcommand{\RENet}{\textsc{RE-Net}\xspace}

\title{Temporal Attribute Prediction via Joint Modeling of Multi-Relational\\ Structure Evolution}

\newcommand{\mat}[1]{\mathbf{#1}}

\newcommand{\nop}[1]{}

\newcommand{\para}[1]{\smallskip\noindent\textbf{#1}}
\newcommand{\prob}{Pr}

\newcommand\Tau{\mathrm{T}}

\makeatletter
\newcommand{\printfnsymbol}[1]{%
  \textsuperscript{\@fnsymbol{#1}}%
}
\makeatother

\author{
Sankalp Garg$^1$\footnote{Equal Contribution}\footnote{Work done while at USC}\and
Navodita Sharma$^2$\printfnsymbol{1}\printfnsymbol{2}\and
Woojeong Jin$^3$\And
Xiang Ren$^3$\\
\affiliations
$^1$Indian Institute of Technology Delhi\\
$^2$Indian Institute of Technology Madras\\
$^3$University of Southern California\\
\emails
\{sankalp2621998, navoditasharma16\}@gmail.com,
\{woojeong.jin, xiangren\}@usc.edu
}

\begin{document}
\maketitle
\begin{abstract}

Time series prediction is an important problem in machine learning. Previous methods for time series prediction did not involve additional information. With a lot of dynamic knowledge graphs available, we can use this additional information to predict the time series better. Recently, there has been a focus on the application of deep representation learning on dynamic graphs. These methods predict the structure of the graph by reasoning over the interactions in the graph at previous time steps. In this paper, we propose a new framework to incorporate the information from dynamic knowledge graphs for time series prediction. We show that if the information contained in the graph and the time series data are closely related, then this inter-dependence can be used to predict the time series with improved accuracy. Our framework, \frameworkName, learns a static embedding for every node in the graph as well as a dynamic embedding which is dependent on the dynamic attribute value (time-series). Then it captures the information from the neighborhood by taking a relation specific mean and encodes the history information using RNN. We jointly train the model link prediction and attribute prediction. We evaluate our method on five specially curated datasets for this problem and show a consistent improvement in time series prediction results. We release the data and code of model \frameworkName for future research\footnote{\url{https://github.com/INK-USC/DArtNet}}.
\end{abstract}
\section{Introduction}
\begin{figure}[t]
    \centering
    \includegraphics[width=1.0\linewidth]{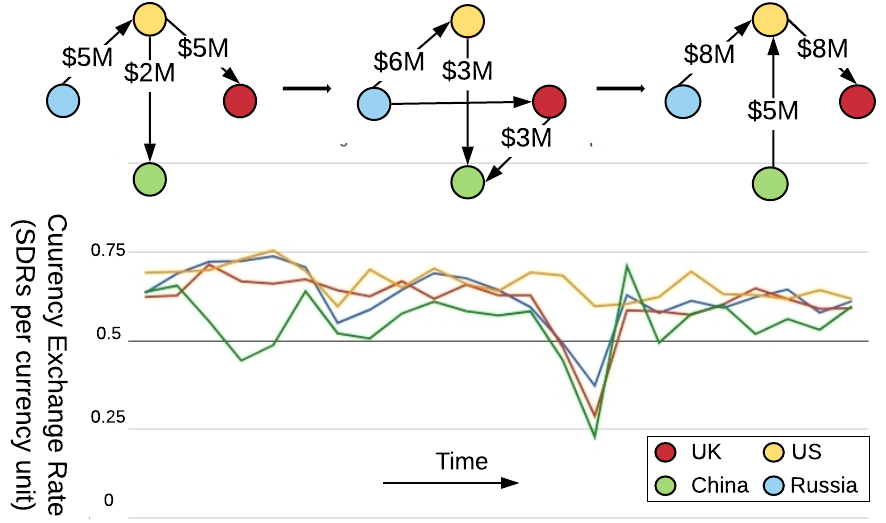}
    \caption{Evolving graph structure and attribute values with time in the Dynamic Attributed Graph. The node represents a country and the edges between them represents the value of trade. The graph shows the changing attribute values (currency exchange rate) for each node.}
    \label{fig:tradegraph}
\end{figure}

Many real-world scenarios exist where the time series can effectively be complemented with external knowledge. One such scenario represents information about the trade between countries in the form of a temporally evolving knowledge graph. The information about trade between the countries affects the corresponding currency exchange rate. Using information about trade, we want to better predict the currency exchange rate with high accuracy.

Time Series forecasting deals with the prediction of the data points of the sequence at a future timestamp based on the data available up till the current timestamp. Methods such as Auto-Regressive Integrated Moving Average (ARIMA) model and Kalman filtering \cite{Liu:2011:DSC:2020408.2020571,Lippi2013ShortTermTF} are popular for predicting time series. Representational learning on graph-structured data \cite{RL1,survey1,gcn,graphrnn} is a widely researched field with considerable focus on temporally-evolving graphs \cite{dyrep,dynamicApp1,dynamicApp2,dynamicApp3}. The increasing amount of complex data that can be effectively represented using dynamic multi-relational graphs \cite{kg1} has led to this increased focus on dynamic graph modeling. Several methods such as ConvE~\cite{convE}, RGCN~\cite{RGCN}, and DistMult~\cite{DistMult} have shown admirable results on modeling static, multi-relational graph data for link prediction. There are other approaches that attempt to model dynamic knowledge graphs by incorporating temporal information, and these include Know-Evolve~\cite{know-evolve}, HyTE~\cite{hyte}, and TA-DistMult~\cite{TA-distmult}, among others. 

The two aforementioned fields of time series prediction and representation learning on graphs have mainly been separated in the machine learning community. Recently some work has been done in integrating the two fields \cite{DCRNN-YanLiu}, which describes a method to incorporate a static, uni-relational graph for traffic flow prediction. However, this method is only limited to static graphs with a single relation. To date, no method has been proposed for integrating temporally evolving graphs and time series prediction. In this paper, we propose a new method for exploiting the information from the dynamic graphs for time series prediction. We propose the use of static learnable embedding to capture the spatial information from knowledge graphs and a dynamic embedding to capture the dynamics of the time series and the evolving graph.

We present the first-ever solution to the problem of time-series prediction with temporal knowledge graphs (TKG). Since, to the best of our knowledge, currently no datasets exist which align with the problem statement, we prepared five suitable datasets through web scraping and evaluate our model. We show that our approach beats the current state-of-the-art methods for time series forecasting on all the five datasets. Our approach also predicts the time series by any number of time steps and does not require a test time graph structure for evaluation. 

\section{Related Work}
We review work using static graphs for time series prediction and work on temporal knowledge graphs.


\paragraph{Time-series Prediction.}
In addition to the general time-series prediction task, there have been some recent studies on the spatial-temporal forecasting problem.
Diffusion Convolutional Recurrent Neural Network (DCRNN) \cite{DCRNN-YanLiu} is a method which incorporates a static, uni-relational graph for time series (traffic flow) forecasting. Traffic flow is modeled as a diffusion process on a directed graph. The method makes use of bidirectional random walks on the graph to capture the spatial dependency and uses an encoder-decoder framework with scheduled sampling for incorporating the temporal dependence. However, this method cannot be extended to temporally evolving graphs as well as multi-relational graphs. Another paper on Relational Time Series Forecasting \cite{rossi_2018}, also formulates the problem of using dynamic graphs for time series prediction though it is not formulated for multi relational data. Neural relational inference \cite{kg2} also looks at the inverse problem of predicting dynamics of graph with attribute information.

\paragraph{Temporal Knowledge Graph Reasoning and Link Prediction.}
There have been several attempts on reasoning on dynamically evolving graphs. HyTE \cite{hyte} is a method for embedding knowledge graphs which views each timestamp in the graph data as a hyperplane. Each {head, relation, tail} triple at a particular timestamp is projected into the corresponding hyperplane. The translational distance, as defined by the TransE model \cite{transE}, of the projected embedding vectors, is minimized. 
TA-Distmult is a temporal-aware version of Distmult. For a quadruple $h,r,t,\tau$, a predicate $p$ is constructed using $r, \tau$ which is passed into an GRU. The last hidden state of the GRU is taken as the representation of the predicate sequence ($e_p$). 
Know-Evolve~\cite{know-evolve} models a relationship between two nodes as a multivariate point process. Learned entity embeddings are used to calculate the score for that relation, which is used to modulate the intensity function of the point process. ReNet \cite{renet} uses the neighborhood aggregation and RNN to capture the spatial and temporal information in the graph. \par


\begin{figure}[t]
    \centering
    \includegraphics[width=\linewidth]{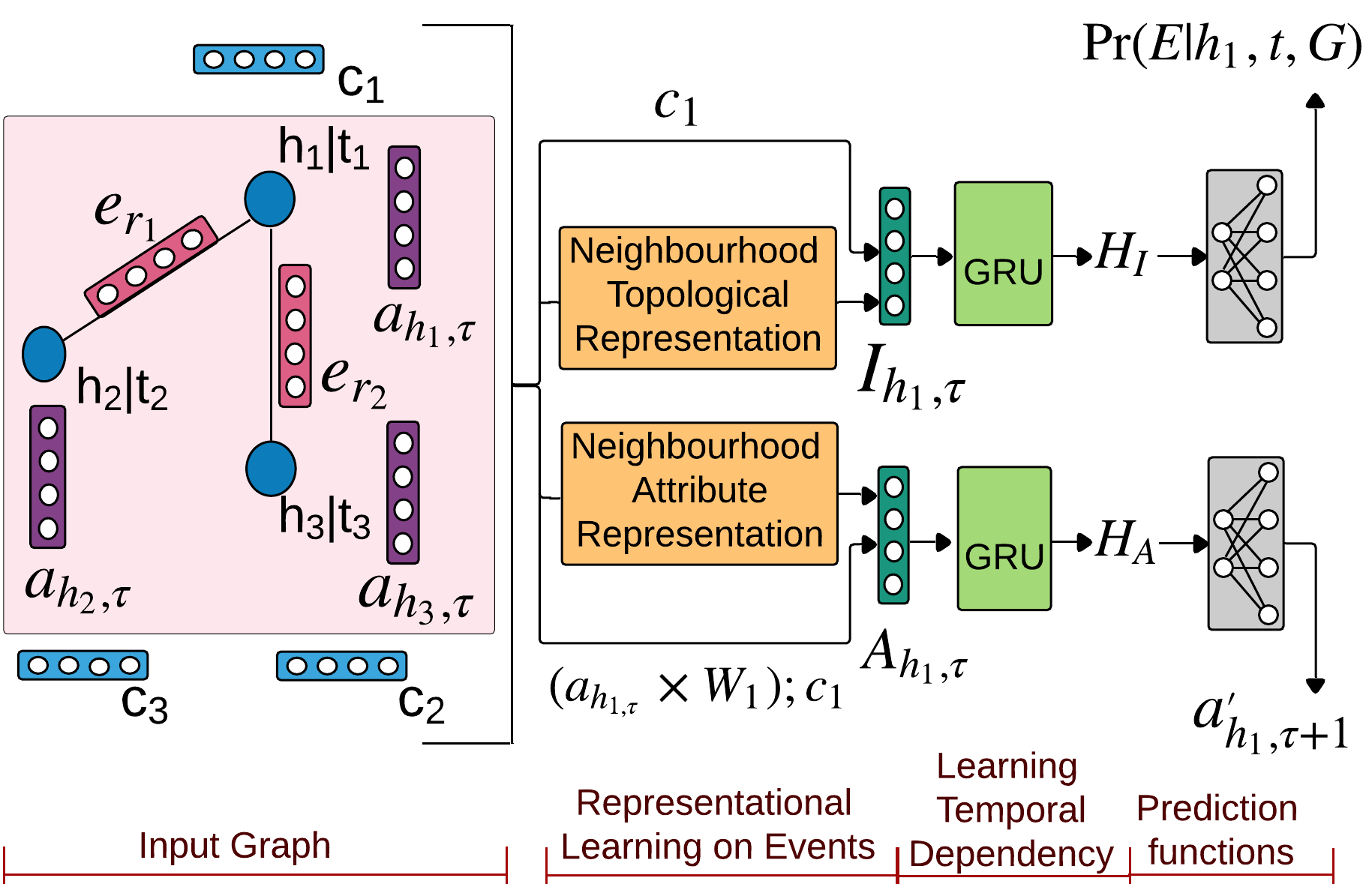}
    \caption{Network for Link Prediction and Attribute Prediction. $h_i$ and $t_i$ denote the $i^{th}$ head and tail respectively. $e_{r_i}$ denotes the embedding for the relation $r_i$ and $a_{h_i,\tau}$ denotes the attribute values associated with $h_i$ at time $\tau$. From neighbouring nodes, topological representation is obtained and using attribute values, attribute representation is obtained which are then fed to recurrent network. The recurrent network's output is used to obtain the predictions.} 
    \label{fig:dan1}
\end{figure}

\section{Problem Formulation}
\label{Problem formulation}
A Knowledge Graph is a multi-relational graph that can be represented in the form of triples $(h,r,t)$ where $h$ denotes the head, $t$ denotes the tail, and $r$ is the relation between the nodes $h$ and $t$. A TKG has a time dimension as well, and the graph can be represented as quadruples in the form $(h,r,t,\tau)$ where $\tau$ denotes the timestamp at which the relation $r$ exists between the nodes $h$ and $t$. 
We now introduce Dynamic Attributed Graphs, formalize our problem statement, and in later sections, present our model for making predictions on Dynamic Attributed Graphs.

\paragraph{Problem Definition.} 
A Dynamic Attributed Graph (DAG) is a directed graph, where the edges are multi-relational with time stamp associated with each edge known as an event, and attributes associated with the nodes for that particular time. An event in dynamic graph is represented as a quadruple $(h,r,t,\tau)$ where $h$ is the head entity, $r$ is the relation, $t$ is the tail entity and $\tau$ is the time stamp of the event. In a dynamic attributed graph, an event $(E_\tau)$ is represented as a hextuple $(h,r,t,a^\tau_h,a^\tau_t,\tau)$ where $a^\tau_h$ and $a^\tau_t$ are the attribute associated with head and tail at time $\tau$. The collection of all the events at a time constitutes a dynamic graph $G_\tau$ where $G_\tau = \{E_{\tau_i} = (h_i,r_i,t_i,a^\tau_{h_i},a^\tau_{t_i},\tau_i) | \tau_i=\tau, \forall i\}$. 
The goal of the DAG Prediction problem is to learn a representation of the dynamic graph events and predict the attributes at each node at future timestamps by learning a set of functions to predict the events for the next time step. Link is predicted jointly, to aid the attribute prediction task. Formally, we want to learn a set of functions $\{F(.)\}$ such that:
\begin{equation*}
    [G_1,G_2,\dots,G_\tau]\xrightarrow{\{F(.)\}}[G_{\tau+1},G_{\tau+2},\dots,G_{\tau+\Tau}].
\end{equation*}

We divide the dynamic graph at any time $\tau$ in two sets, one consisting of only $(h,r,t,\tau)$ and other consisting of only the attribute values. Formally, $G_\tau = G^I_{\tau}\cup G^A_{\tau}$, where $G^I_{\tau} = \{E^I_{\tau_i} = (h_i,r_i,t_i,\tau_i) | \tau_i=\tau, \forall i\}$ and $G^A_{\tau} = \{a^{\tau}_{h_i} | \tau_i=\tau, \forall i\}$. 
We propose to predict $G^I$ and $G^A$ using these set of functions as follows:
\begin{equation*}
    [G_1,G_2,\dots,G_\tau]\xrightarrow{\{F(.)\}}[(G^I_{\tau+1}, G^A_{\tau+1}),\dots,(G^I_{\tau+\Tau}, G^A_{\tau+\Tau})].
\end{equation*}
We jointly predict the graph structure $G^I_{\tau'}$ and the attribute values $G^A_{\tau'}$ and show that the attribute values are being predicted with greater accuracy than can be done using any existing method.

\section{Proposed Framework : \frameworkName}
\label{Framework}
We now present our framework for learning the set of functions for predicting the events for the next timestamp, given the history. We name our framework \textbf{\frameworkName}  which stands for \textbf{D}ynamic \textbf{A}ttributed \textbf{N}etwork.\par
We model the changing graph structure and attribute values by learning an entity-specific representation. We define the representation of an entity in the graph as a combination of a static learnable embedding which does not change with time and represent static characteristics of each node, and a dynamic embedding which depends on the attribute value at that time and represent dynamically evolving property of each node. We then aggregate the information using the mean of neighborhood entities. For every entity in the graph, we model history encoding using a Recurrent Neural Network. Finally, we use a fully connected network for attribute prediction and link prediction tasks. 
\subsection{\textbf{Representation Learning on Events}}
The main component of our framework is to learn a representation over the events which will be used for predicting future events. We learn a head-specific representation and then model its history using Recurrent Neural Network (RNN). Let $E_{h,\tau}$ represent the events associated with head $h$ at time $\tau$, i.e., $E_{h,\tau} = \{(h,r_i,t_i,a^\tau_{h},a^\tau_{t_i},\tau_i) | \tau_i=\tau, \forall i\}$. For each entity in the graph, we decompose the information in two parts, static information and dynamic information. Static information does not change over time and represents the inherent information for the entity. Dynamic information changes over time. It represents the information that is affected by all the external variables for the entity. For every entity $h$ in the graph at time $\tau$, we construct an embedding for the entity which consists of two components:
\begin{enumerate}
    \item Static (Constant) learnable embedding $(\mat{c}_h)$ which does not change over time.
    \item Dynamic embedding $(\mat{d}_{h,\tau} = \mat{a}_{h,\tau}\cdot\mat{W}_1)$ which changes over time.
\end{enumerate}
where $\mat{c}_h \in \mathbb{R}^{d}$, $a_{h,\tau} \in \mathbb{R}^k $ is the attribute of entity $h$ at time $\tau$  and $\mat{W}_1 \in \mathbb{R}^{k\times d}$ is learnable parameter. The attribute value can be a multi-dimensional vector with $k\geq1$ representing multiple time-series associated with same head. Hence the embedding of entity $h$ becomes $\mat{e}_{h,\tau} = (\mat{c}_h;\mat{d}_{h,\tau})$ where $;$ is the concatenation operator. For every relation (link) $r$, we construct a learnable static embedding $\mat{e}_r \in \mathbb{R}^{d}$.\par

To capture the neighbourhood information of entity $h$ at time $\tau$ from the dynamic graph, we propose two spatial embeddings: attribute embedding $(\mat{A}_{h,\tau})$ and interaction embedding $(\mat{I}_{h,\tau})$. $\mat{A}_{h,\tau}$ captures the spatio-attribute information from the neighbourhood of the entity $h$ and $\mat{I}_{h,\tau}$ captures the spatio-interaction information from the neighbourhood of the entity $h$. Mathematically, we can define the spatial embeddings as:
\begin{equation}
    \mat{A}_{h,\tau} = (\mat{e}_{h,\tau};\frac{1}{|E_{h,\tau}|}\mathlarger{\mathlarger{\Sigma}}_{(h,r_j,t_j,a^\tau_{h},a^\tau_{t_j},\tau)\in E_{h,\tau}}{(\mat{e}_{t_j,\tau};\mat{e}_{r_j})\cdot\mat{W}_2)},
\end{equation}
\begin{equation}
    \mat{I}_{h,\tau} = (\mat{c}_{h};\frac{1}{|E_{h,\tau}|}\mathlarger{\mathlarger{\Sigma}}_{(h,r_j,t_j,a^\tau_{h},a^\tau_{t_j},\tau)\in E_{h,\tau}}{(\mat{c}_{t_j};\mat{e}_{r_j})\cdot\mat{W}_3)},
\end{equation}
where $|E_{h,\tau}|$ is the cardinality of set $E_{h,\tau}$ and $\mat{W}_2\in \mathbb{R}^{3d\times d}$ and $\mat{W}_3\in \mathbb{R}^{2d\times d}$ are learnable parameters.

\subsection{Learning Temporal Dependency}
The embeddings $\mat{I}_{h,\tau}$, $\mat{A}_{h,\tau}$ capture the spatial information for entity $h$ at time $\tau$. For predicting the information at future time, we need to capture the temporal dependence of the information. To keep track of the interactions and the attribute evolution over time, we model the history using Gated Recurrent Unit \cite{GRU}, an RNN. For the head s, we define the encoded attribute history at time $\tau$ as the sequence $[{A}_{h,1}, {A}_{h,2}, \dots , {A}_{h,\tau-1}]$ and the encoded interaction history at time $\tau$ as the sequence $[{I}_{h,1}, {I}_{h,2}, \dots , {I}_{h,\tau-1}]$. These sequence provide the full information about the evolution of the head $h$ till time $\tau$. To represent in sequences, we model the encoded attribute and encoded interaction history for head $h$ as follows:
\begin{align}
    \mat{H}_A(h,\tau) &= GRU^1(\mat{A}_{h,\tau}, \mat{H}_A(h,\tau-1)),\\
    \mat{H}_I(h,r,\tau) &= GRU^2((\mat{I}_{h,\tau}; \mat{c}_h;\mat{e}_{r_j}), \mat{H}_I(h,r,\tau-1)),
\end{align}
where the vector $\mat{H}_A(h,\tau)$ captures the spatio-temporal information for the attribute evolution i.e. captures how the attribute value of the entity evolves over time with respect to the evolving graph structure, while the vector $\mat{H}_I(h,r,\tau)$ captures the spatio-temporal information of how the relation $r$ is associated with the entity $h$ over time. We show the \frameworkName, its input and output in Figure \ref{fig:dan1}.

\begin{table*} [t]

	\centering
	\begin{tabular}{ccccccc}
	    \toprule
	    \textbf{Dataset} & \textbf{\# Train} & \textbf{\# Valid} & \textbf{\# Test} & \textbf{\# Nodes} & \textbf{\# Rel} & \textbf{\# Granularity} \\
	    \midrule
	    AGT & 463,188 & 57,898 & 57,900 & 58 & 178 & Monthly\\
	    CAC(small) & 2070 & 388 & 508 & 90 & 1 & Yearly \\
	    CAC(large) & 116,933 & 167,047 & 334,096 & 20,000 & 1 & Yearly \\
	    MTG & 270,362 & 39,654 & 74,730 & 44 & 90 & Monthly\\
	    AGG & 3,879,878 & 554,268 & 1,108,538 & 6,635 & 246 & Monthly\\
	    
	    \bottomrule
    \end{tabular}
    \caption{Statistics of the five datasets used in the experiments.}
    \label{tab::data}
\end{table*}

\subsection{\textbf{Prediction Functions}} The main aim of the model is to be able to predict the future attribute values as well as the interaction events. To get the complete information of the event at next time step, we perform the prediction in two steps: (1) prediction of the attribute values for the whole graph and (2) prediction of the interaction events for the graph. We know $G_{\tau+1} = \{E_{\tau_i} = (h_i,r_i,t_i,a^\tau_{h_i},a^\tau_{t_i},\tau_i) | \tau_i=\tau+1, \forall i\}$. To predict $G_{\tau+1}$, we divide it into two sets $G^I_{\tau+1}$ and $G^A_{\tau+1}$. The attribute values of $G^A_{\tau+1}$ are predicted directly and modelled as follows:
\begin{equation}
    a'_{h,\tau+1} = f_A(\mat{H}_A(h,\tau), \mat{c}_h).
\end{equation}
The attribute value for the entity is a function of the spatio-attribute history for the entity and the static information about the entity. 
Attribute prediction requires graph structures, so we also predict graph structures. The probability of $G^I_{\tau+1}$ is modeled as:
\begin{align*}
    \prob(G^I_{\tau+1}|G_{1},\dots,G_{\tau}) &= \prod_{E^I_k \in G^I_{\tau+1}} \prob(E^I_k|G_{1},\dots,G_{\tau}).
\end{align*}
At $\tau+1$, we can write this probability as
\begin{multline*}
\prob(E^I_k|G_{1},\dots,G_{\tau}) = \prob(h_k,r_k,t_k|G_{1},\dots,G_{\tau})\\
= \prob(t_k|h_k,r_k,G_{1},\dots,G_{\tau})\cdot\prob(h_k,r_k|G_{1},\dots,G_{\tau}). 
\end{multline*}
In this work, we consider the case that probability of $(h,r)$ is independent of the past graphs $\{G_{1},\dots,G_{\tau}\}$, and model it using uniform distribution, leading to
\begin{equation*}
    \prob(G^I_{\tau+1}|G_{1},\dots,G_{\tau}) \propto \prod_{E^I_k \in G^I_{\tau+1}} \prob(t_k|h_k,r_k,G_{1},\dots,G_{\tau}).
\end{equation*}
For predicting the interaction at future timestamp, we model the probability of the tail as follows:
\begin{equation}
    \prob(t|h,r,G_{1},\dots,G_{\tau}) = f_I(\mat{H}_I(h,r,\tau), \mat{c}_h, \mat{e}_r).
\end{equation}
The functions $f_A$ and $f_I$ can be any function. In our experiments we use the functions as a single-layered feed-forward network.

\subsection{\textbf{Parameter Learning}}
We use multi-task learning \cite{MTLSurvey,MTLRich} loss for optimizing the parameters. We minimize the attribute prediction loss and graph prediction loss jointly. The total loss $L = L_I+\lambda L_A$, where $L_I$ is interaction loss, $L_A$ is attribute loss and $\lambda$ is a hyperparameter deciding the weight of both the tasks. For the attribute loss, we use the mean-squared error $L_A = \frac{1}{N}\sum_{i=1}^{N}(\mat{a'}_{h_i,\tau_i}-\mat{a}_{h_i,\tau_i})^2$, where $a'_{h_i,\tau_i}$ is the predicted attribute and $a_{h_i,\tau_i}$ is the ground truth attribute. For the interaction loss, we use the standard multi-class cross entropy loss, $ L_I =\sum_{i=1}^{N} \sum_{c=1}^{M} y_j\log(\prob(t=c|h_i,r_i))$, where the $M$ is the number classes i.e. number of entities in our case.

\subsection{\textbf{Forecasting Over Time}}
At each inference time, \frameworkName predicts future interactions and attributes based on the previous observations, i.e., $[(G_i^I, G_i^A)]_{i=0}^{\tau}$.
To predict interactions and attributes at time $\tau+\Delta \tau$, \frameworkName adopts multi-step inference and predicts in a sequential manner.
At each time step, we compute the probability $\prob(t|h,r,G_{1},\dots,G_{\tau})$ to predict $G_{\tau+1}^A$. We rank the tails predicted and choose the top-$k$ tails as the predicted values. We use the predicted tails as $G_{\tau+1}^A$ for further inference. Also, we predict attributes, which yields $G_{\tau+1}^I$. Now we have graph structure and attributes at time $\tau+1$. We repeat this process until we get $(G_{\tau+\Delta \tau -1}^I, G_{\tau+\Delta \tau -1}^A)$. Then we can predict interactions and attributes at time $\tau+\Delta \tau$ based on $[(G_i^I, G_i^A)]_{i=0}^{\tau+\Delta \tau -1}$.

\begin{table*}[t]

\centering

\scalebox{0.95}{
\begin{tabular}{llccccc}
\toprule
&\textbf{Method}                       & \textbf{ATG} &\textbf{CAC(small)}&\textbf{ CAC(large)} & \textbf{MTG} & \textbf{AGG} \\                      
& & \textbf{($10^{-3}$)} &\textbf{($10^{-2}$)}&\textbf{($10^{-4}$)} & \textbf{($10^{-4}$)} & \textbf{($10^{-4}$)} \\
\cmidrule(lr){2-7}
\multirow{4}{*}{\rotatebox{90}{\hspace*{-6pt}w/o graph}} 
       & Historic Average                                                                                                & 1.636 & 4.540      & 9.810      & 14.930 & 600.000 \\
       & VAR\setfootnotemark\label{first}\cite{VAR}                     & 3.961 & 6.423      & 10.330     & 9.490  & 300.000 \\
       & ARIMA                                                                                                           & 1.463 & 4.245      & 9.102      & 2.860  & 51.240  \\
       & Seq2Seq model \cite{seq2seq}                                                                   & 1.323 & 4.554      & 8.080      & 2.975  & 28.000  \\

\cmidrule(lr){2-7}
\multirow{11}{*}{\rotatebox{90}{\hspace*{-6pt}with graph}} 
       & ConvE+GRU (1 layer) \setfootnotemark\label{third}\cite{convE} & 0.763 & 3.899      & 8.220      & 7.240  & 202.580 \\
       & ConvE+GRU (2 layers)                                                                                           & 0.728 & 4.321      & 8.440      & 9.460  & 206.640 \\
       & HyTE+GRU (1 layer) \setfootnotemark\label{second}\cite{hyte}  & 4.041 & 40.234     & 8.089      & 37.170 & 7.430   \\
       & HyTE+GRU (2 layers)                                                                                            & 1.531 & 40.885     & 8.230      & 17.410 & 2.070   \\
       & TA-Distmult+GRU (1 layer) \cite{TA-distmult}                                                  & 0.847 & 3.584      & 9.456      & 16.880 & 3.250   \\
       & TA-Distmult+GRU (2 layer)                                                                                      & 0.796 & 3.432      & 9.034      & 9.770  & 7.030   \\

\cmidrule(lr){2-7}
       & RENet (mean)+GRU (1 layer) \cite{renet}                                                       & 0.793 & 4.073      & 9.022      & 5.020  & 203.320 \\
       & RENet (mean)+GRU (2 layers)                                                                                    & 0.857 & 3.865      & 8.856      & 4.348  & 200.220 \\
       & RENet (RGCN)+GRU (1 layer)                                                                                     & 0.620 & 3.718      & 8.998      & 5.170  & 203.120 \\
       & RENet (RGCN)+GRU (2 layers)                                                                                    & 0.550 & 3.984      & 8.201      & 12.700 & 201.560 \\
 
\cmidrule(lr){2-7}
       & \frameworkName                                                                                   & \textbf{0.115} & \textbf{3.423}      & \textbf{7.054 }     & \textbf{0.496}  & \textbf{0.848 }  \\

\bottomrule
\end{tabular}
}
\afterpage{\footnotetext[\getrefnumber{first}]{\url{https://github.com/liyaguang/DCRNN}}
             \footnotetext[\getrefnumber{third}]{\url{https://github.com/TimDettmers/ConvE}}
             \footnotetext[\getrefnumber{second}]{\url{https://github.com/malllabiisc/HyTE}}}
\caption{Performance comparison on Attribute Prediction on various datasets. \frameworkName performs best on all the datasets. Relational baselines generally perform better the non-relational baselines. [Smaller is better]}
\label{AttributePredictionresults}
\end{table*}

\section{Experiments}

In this section, we evaluate our proposed method \frameworkName on a temporal attribute prediction task on five datasets.
The attribute prediction task predicts future attributes for each node.

We evaluate our proposed method on two tasks: (1) predicting future attributes associated with each node on five datasets; 
(2) studying variations and parameter sensitivity of our proposed method. We will summarize the datasets, evaluation metrics, and baseline methods in the following sections.

\subsection{Datasets}
Due to the unavailability of datasets satisfying our problem statement, we curated appropriate datasets by scraping the web. We created and tested our approach on the datasets described below. Statistics of datasets are described in Table~\ref{tab::data}.

\paragraph{Attributed Trade graph (ATG).} This dynamic graph represents the net export from one country (node) to another, where each edge belongs to an order of trade segment (in a million dollars). The month-averaged currency exchange rate of the corresponding country in SDRs per currency unit is the time series attribute value.

\paragraph{Co-authorship-Citation dataset (CAC).} Each edge in the graph represents the collaboration between the authors (node) of the research paper. The number of citations per year for an author is the corresponding time series attribute for the node.

\paragraph{Multi-attributed Trade graph (MTG).} 
This is a subset of ATG, with a multi attributed time series representing monthly Net Export Price Index and the value of International Reserves assets in millions of US dollars.

\paragraph{Attributed GDELT graph (AGG).} Global Database of Events, Language, and Tone (GDELT) represents a different type of event in a month between entities like political leaders, organizations, and countries, etc. Here only country nodes are associated with a time-series attribute, which is taken as the Currency Exchange Rate.

\subsection{Evaluation Metrics}
The aim is to predict attribute values at each node at future timestamps. For this purpose, Mean Squared Error (MSE) loss is used. The lower MSE indicates better performance.

\subsection{Baseline Methods}

We show that the results produced by our model outperform those of the existing time series forecasting models. We compare against two kinds of attribute prediction methods.
\paragraph{Time series prediction without TKG.} These methods do not take into account the graph data and make predictions using just the time-series history available. We compare our model to Historic Average (HA), Vector AutoRegressive (VAR) model \cite{VAR}, Autoregressive Integrated Moving Average (ARIMA) model, and GRU based Seq2Seq \cite{seq2seq}. HA makes predictions based on the weighted average of the previous time series values, ARIMA uses lagged observations for prediction and VAR predicts multiple time series simultaneously by capturing linear interdependencies among multiple time series.
\par
\paragraph{Time series prediction with TKG.} Node embeddings are learned using graph representational learning methods (both static and temporal) like \RENet, HyTE, TA-DistMult, and ConvE. For each node, the attribute value at a particular timestamp is concatenated with the corresponding node embedding, and this data is passed into an GRU network for making predictions.

\paragraph{Hyperparameter settings.} All models are implemented in PyTorch using Adam Optimizer for training. The best hyperparameters are chosen using the validation dataset. Typically increasing value of $\lambda$ gives better results, and the best results on each dataset are reported.

\begin{figure*}[t]
    \centering
    \subfloat[ATG]{\includegraphics[scale=0.25]{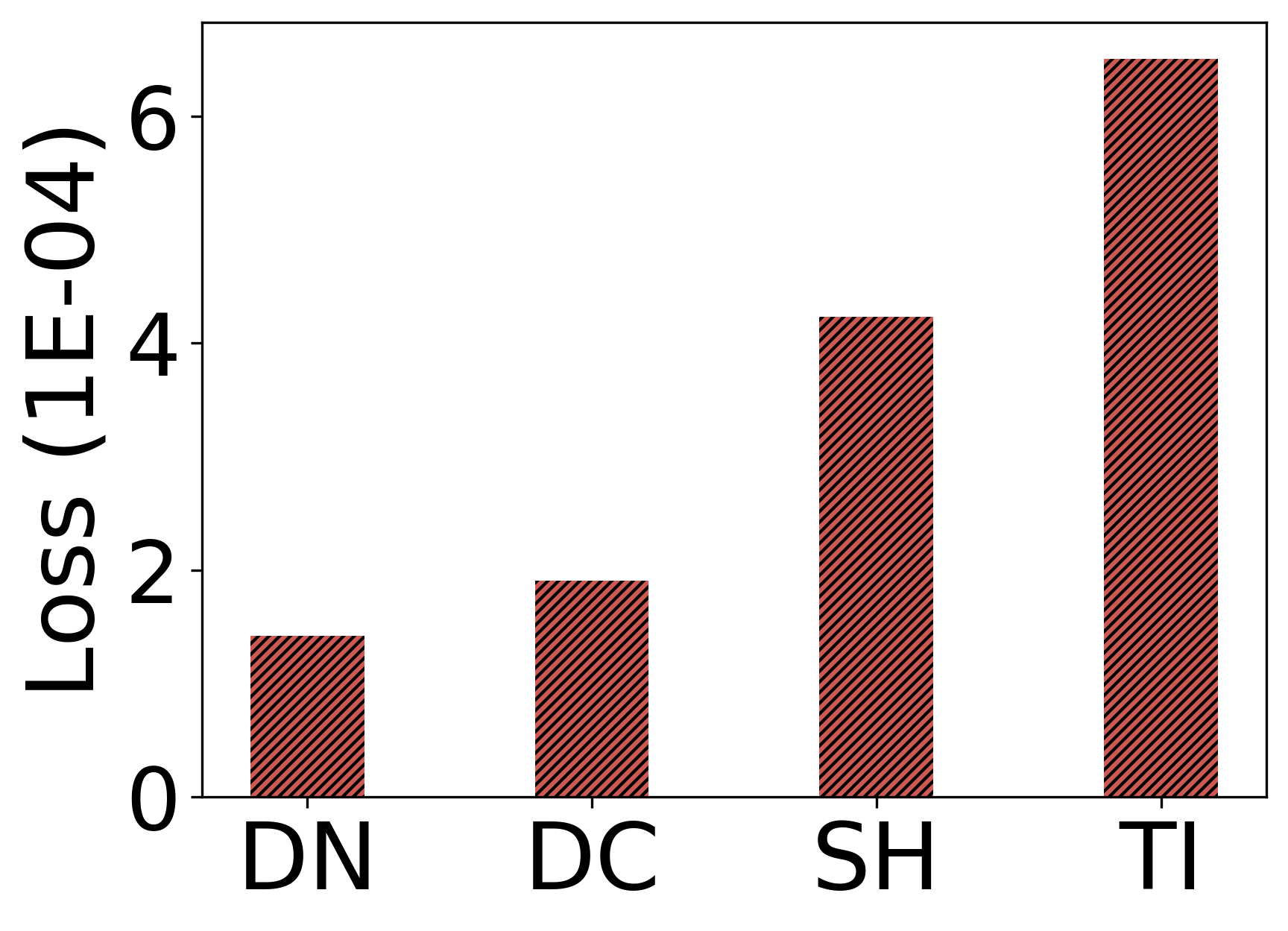}}
    \subfloat[CAC(small)]{\includegraphics[scale=0.25]{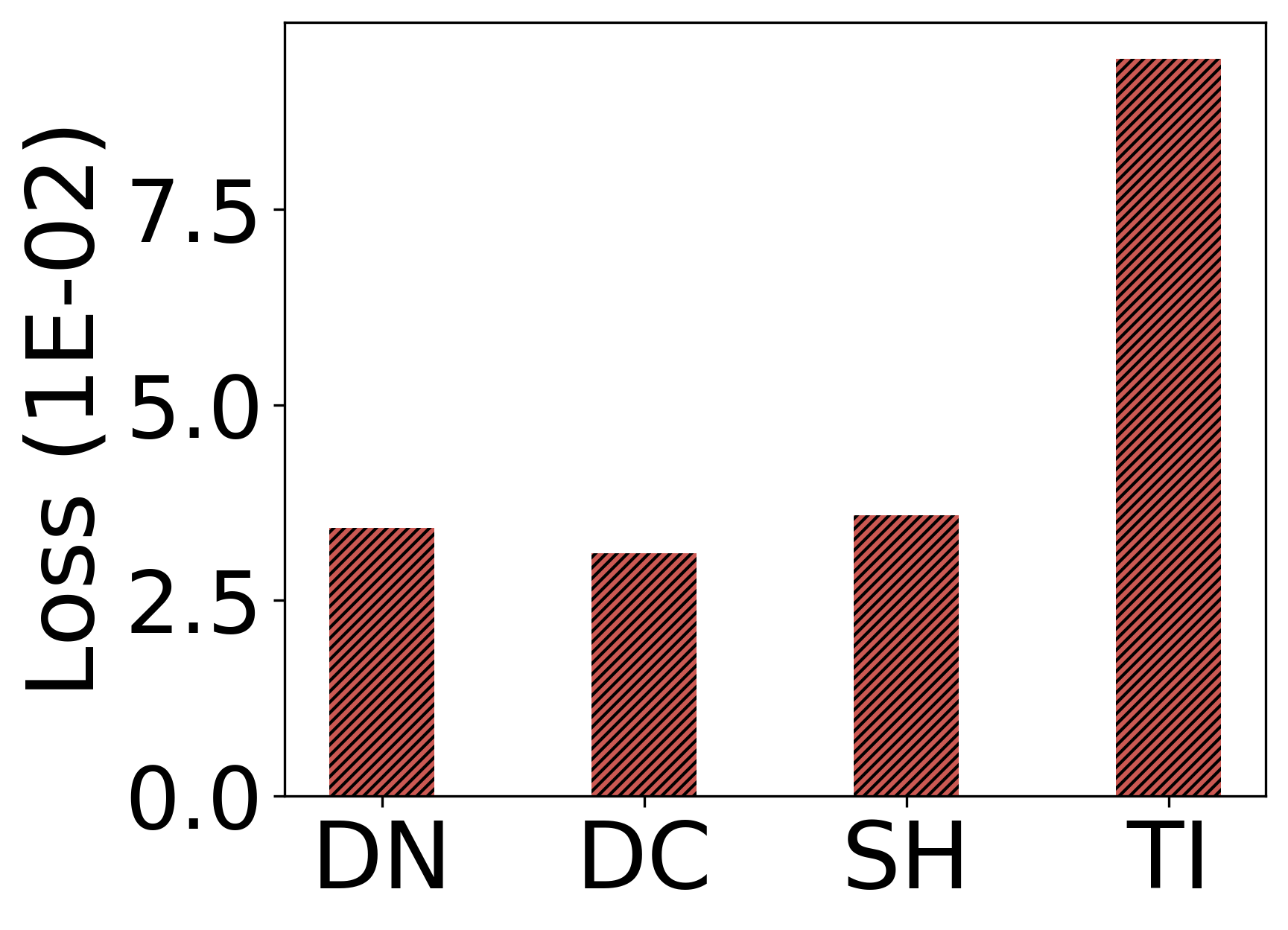}}
    \subfloat[CAC (large)]{\includegraphics[scale=0.25]{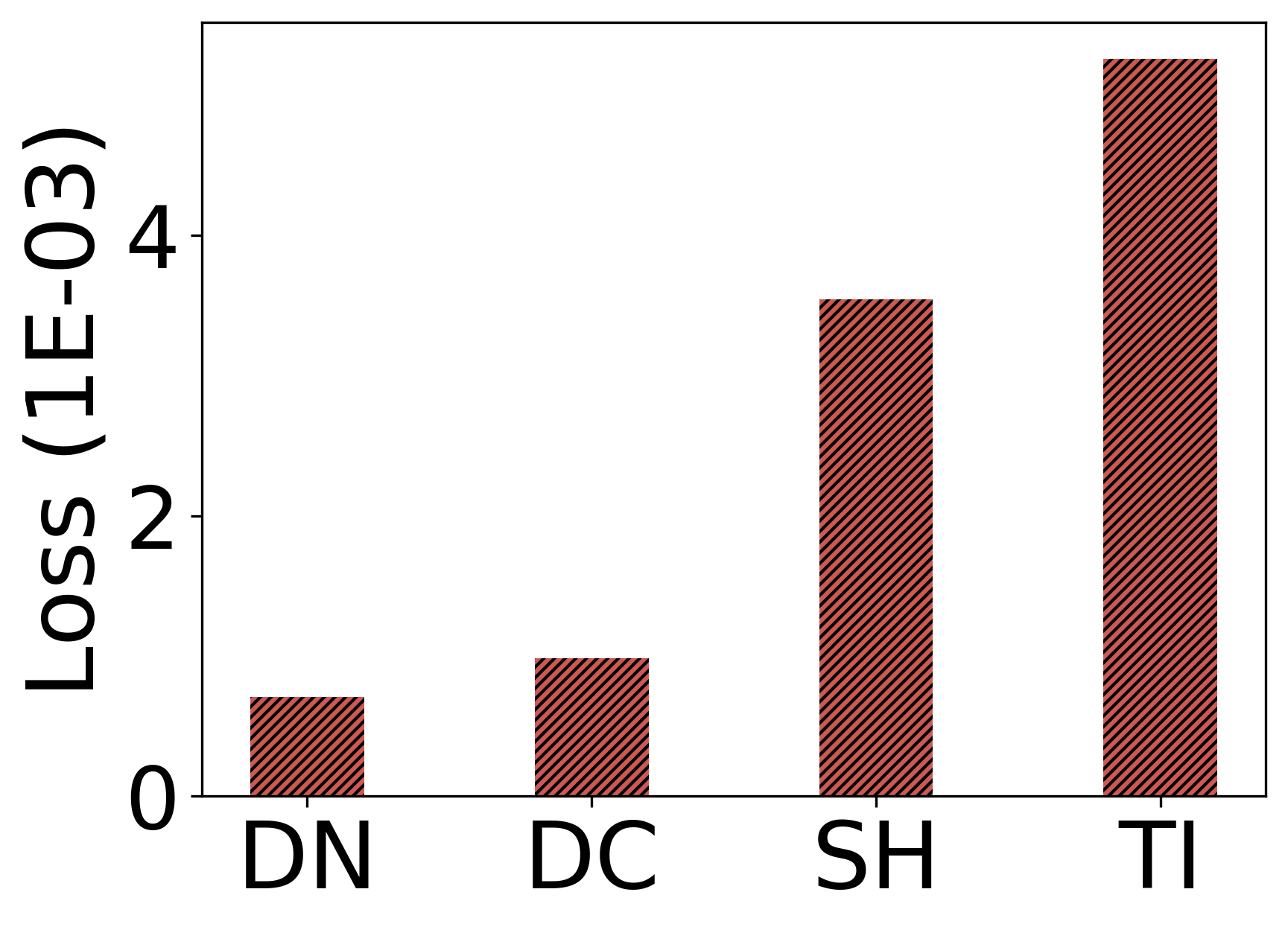}}
    \subfloat[MTG]{\includegraphics[scale=0.25]{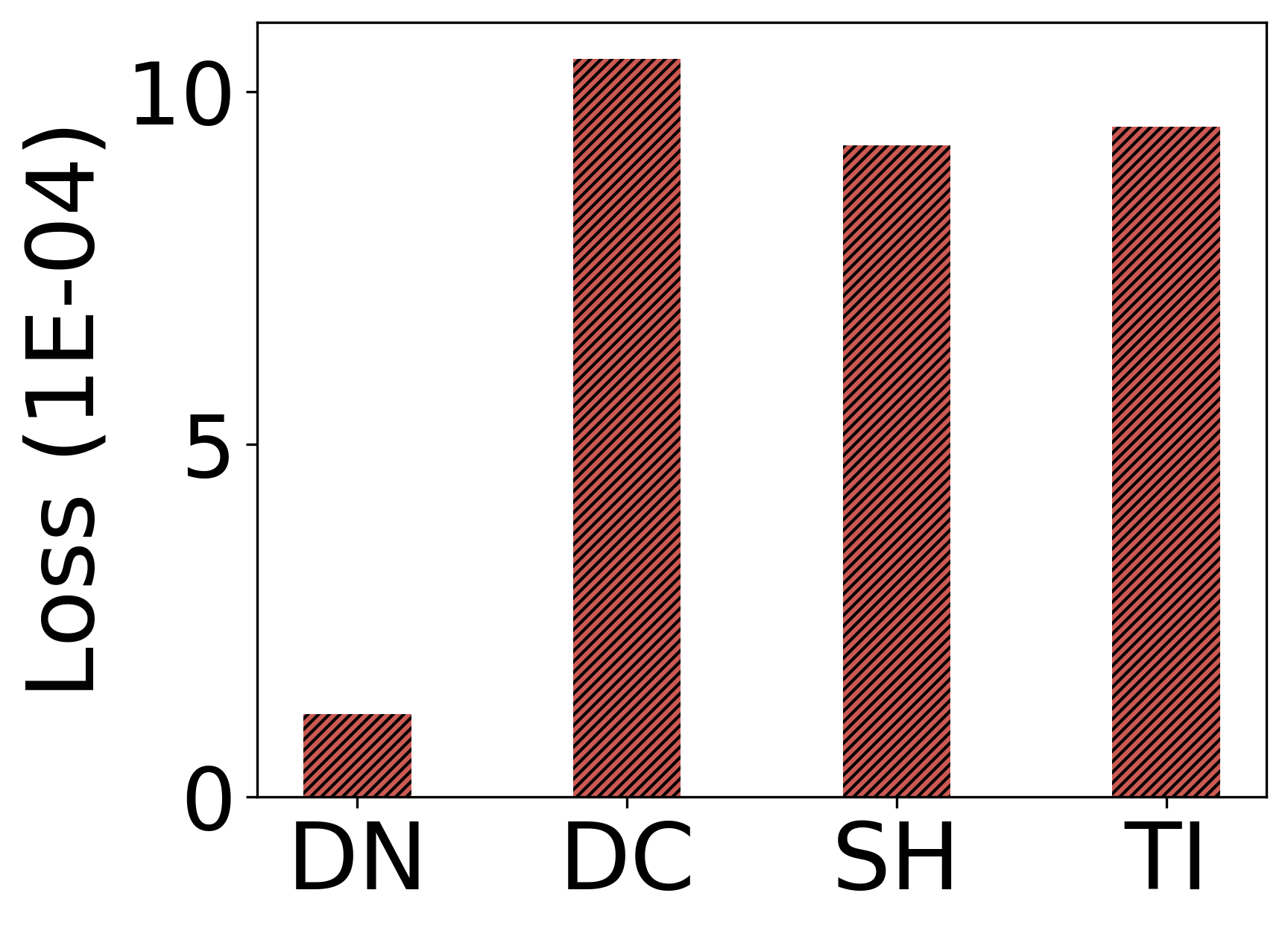}}
    \caption{Variation of Attribute Loss for attribute prediction on ATG, CAC(small), CAC(large), MTG for different variants of \frameworkName [Legend - DN: \frameworkName, DC: Decoupled, SH: Sharing History, TI: Time Independent]. Our method outperforms other methods. We see that our method performs substantially better than decoupled variant, same history setting and the time-independent setting. This suggests that joint modelling of link prediction and attribute prediction is helpful for attribute prediction. It also shows separate history encoding vectors are required for prediction and temporal information is essential for link prediction. [Smaller is better]}
    \label{ablation_loss}
\end{figure*}

\begin{figure*}[t]
    \centering
    \subfloat[ATG]{\includegraphics[scale=0.25]{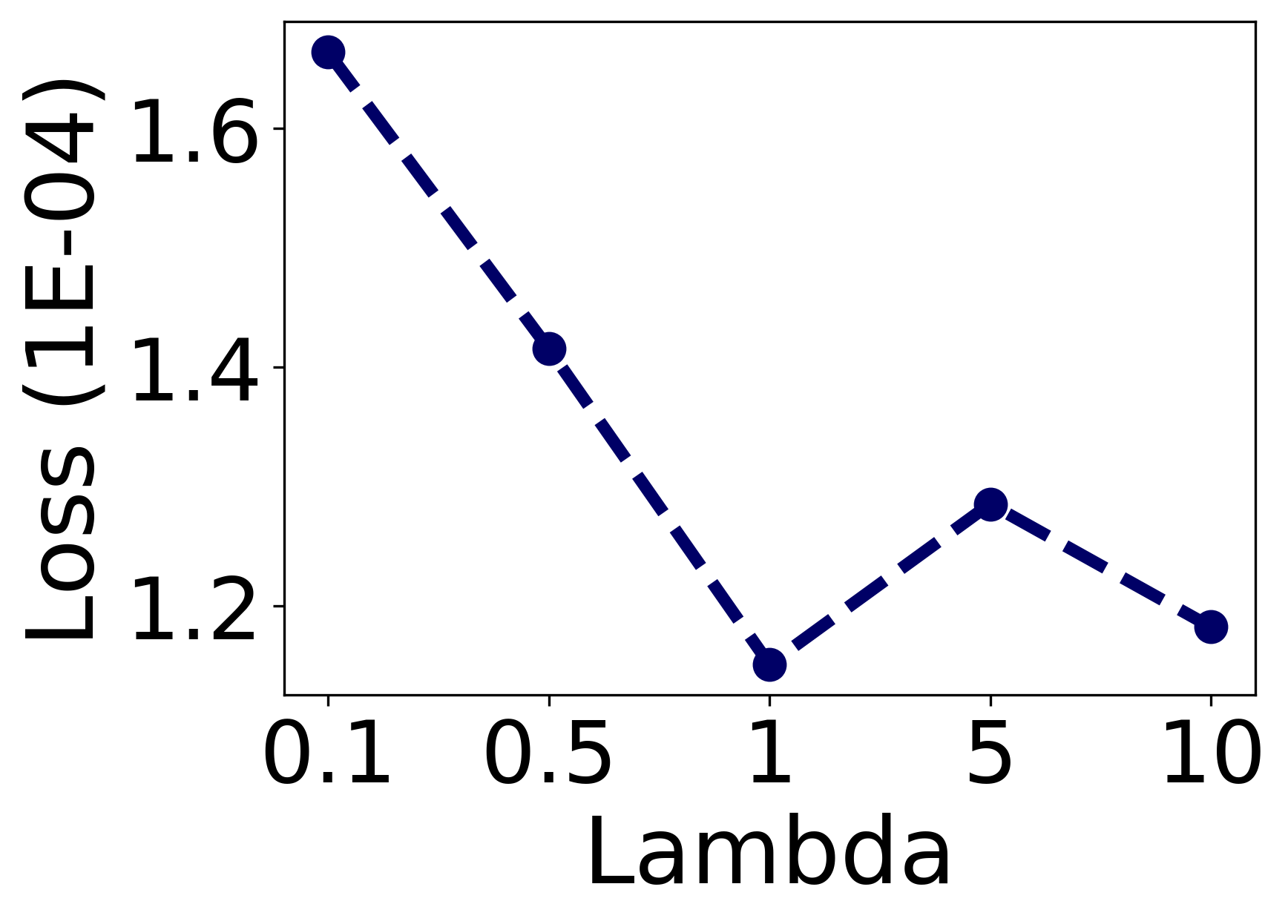}}
    \subfloat[CAC (small)]{\includegraphics[scale=0.25]{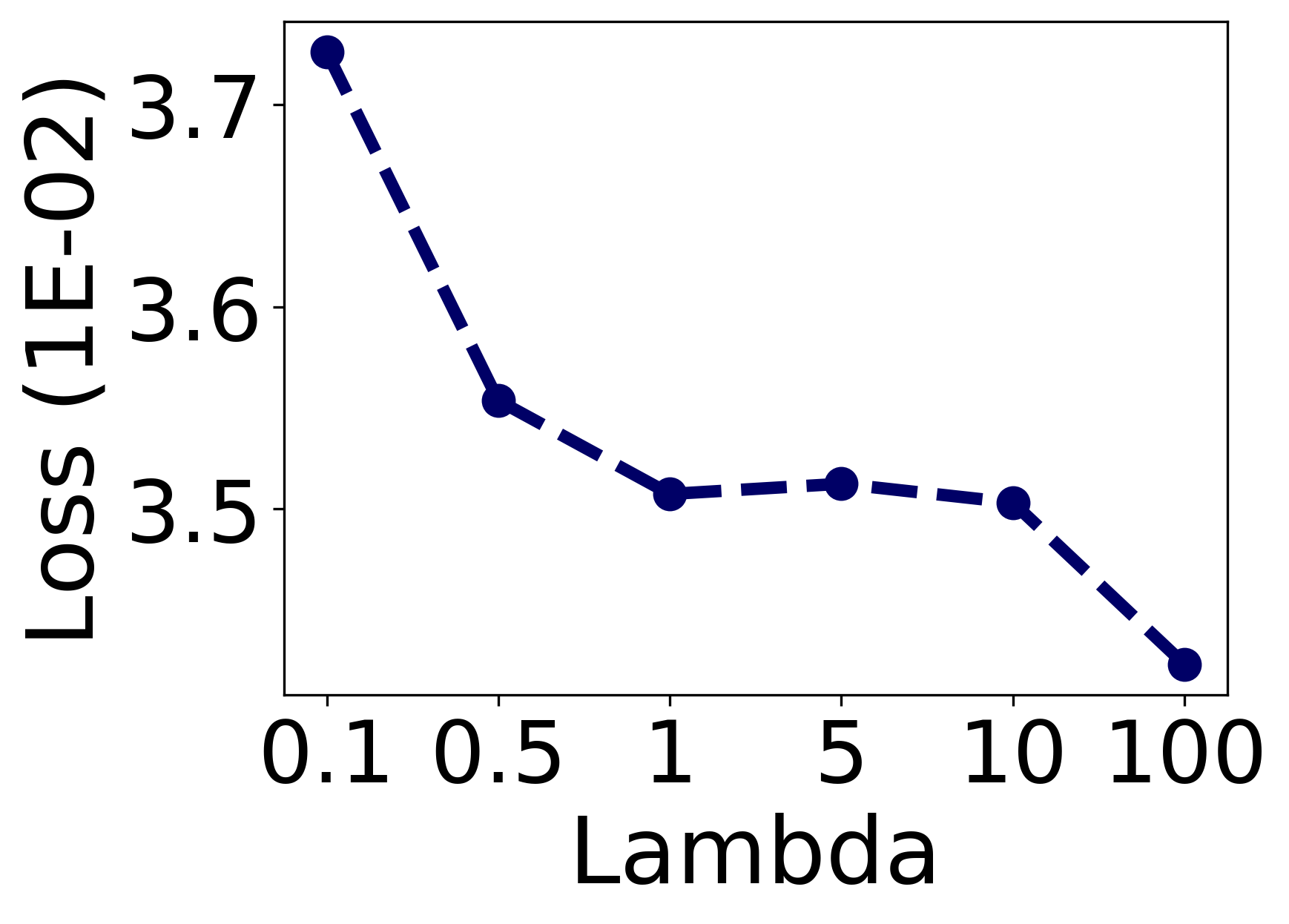}}
    \subfloat[CAC (large)]{\includegraphics[scale=0.25]{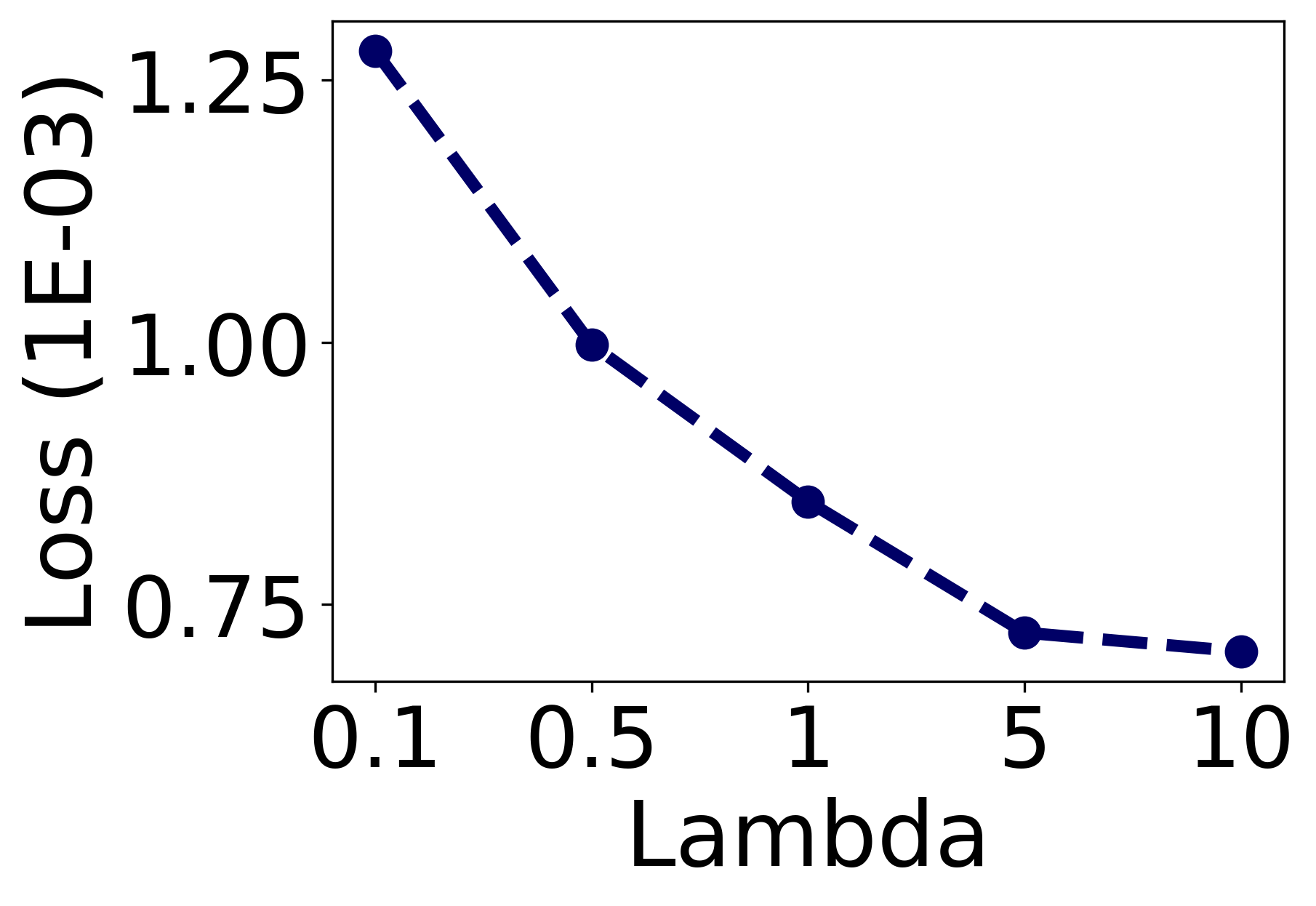}}
    \subfloat[MTG]{\includegraphics[scale=0.25]{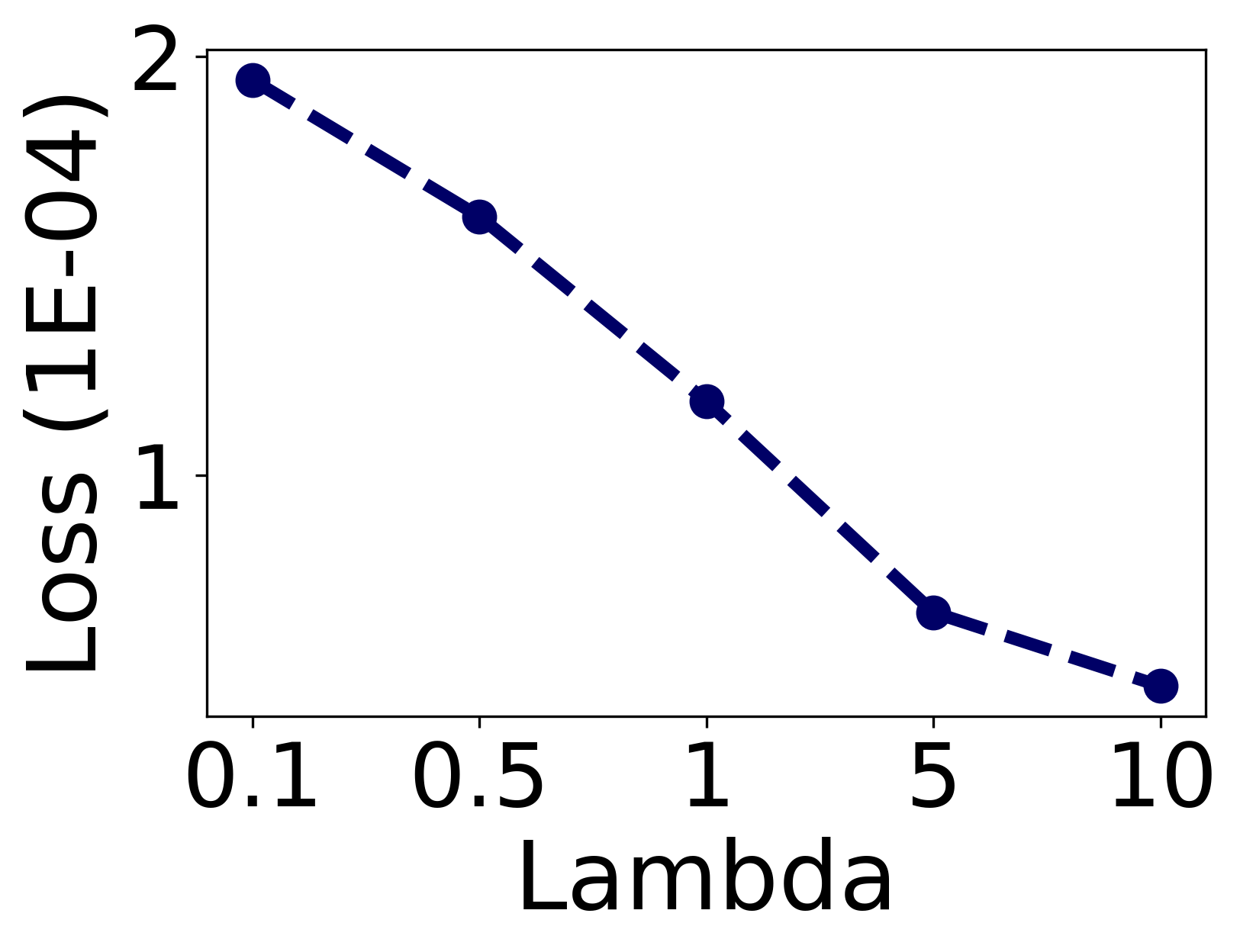}}
    
    \caption{Sensitivity analysis of Attribute Loss with $\lambda$ on ATG, CAC(small), CAC(large), MTG. We see a general trend that the Attribute Loss value decreases with increase in $\lambda$. This is also the expected trend. [Smaller is better]}
    \label{lambad_variation_Loss}
\end{figure*}
\subsection{Main Results}

The results for the attribute prediction on different datasets are reported in Table \ref{AttributePredictionresults}. We see that our method \frameworkName outperforms every other baseline for the attribute prediction by a large margin. From the results it is clear, that the neural network based models outperform the other baselines on these complicated datasets proving their long term modeling capacity. We observe that the relational methods using graph information, generally outperform the non-relational methods on attribute prediction. Large increase in performance is observed for more complicated datasets like MTG and AGG. This suggests that it is the right direction for research to use relational methods for attribute prediction. \frameworkName outperforms other relational methods, which does not jointly train embeddings for attribute prediction and link prediction. This suggests that joint training of embeddings for attribute prediction and link prediction improves the performance on attribute prediction rather than training embeddings separately and then using it for attribute prediction. 

\subsection{Performance Analysis}
\label{Ablation Description}

To study the effects of changes in model parameter sharing and hyperparameter sensitivity on prediction, we perform several ablation studies for \frameworkName on four datasets as AGG does not have attribute values over all nodes. 
\begin{enumerate}

\item{\textbf{Decoupling of Attribute prediction and Interaction prediction tasks.}} We decouple the shared parameters $c_h$ and $\mat{W}_1$ for both tasks and observe the performance. More formally, we use a different embedding for both tasks, i.e $\mat{c}^I_h, \mat{W}^I_1$ and $\mat{c}^A_h, \mat{W}^A_1$ as the parameters for link prediction and attribute prediction task respectively.

\item{\textbf{Sharing history.}} We study the effect of using the same history embedding for both link prediction and attribute prediction. This will help us study if similar history information is required for both the tasks. Here the $H_I$ does not explicitly get the related information so that we can share the weights. Hence the new equations become:
\begin{equation*}
    \mat{H}_I(h,r,\tau) = \mat{H}_A(h,\tau),
\end{equation*}
where the parameters of both the RNNs are shared.

\item{\textbf{Study of Time-Dependent Information.}}
We evaluate the performance of our model in the absence of any temporal information. Hence we do not encode any history for any task and directly predict the tails and the attribute values at a future timestamp. Hence equations are:
\begin{equation*}
    \mat{a}'_{h,\tau+1} = f_A(\mat{c}_h),
\end{equation*}
\begin{equation*}
    P(t|h,r,G_{1},\dots,G_{\tau}) = f_I(\mat{c}_h, \mat{l}_r).
\end{equation*}
\end{enumerate}
\paragraph{Analysis on Variants of \frameworkName.}
Figure \ref{ablation_loss} shows the variation of Attribute Loss with different variants of \frameworkName proposed in Section \ref{Ablation Description}. From Figure \ref{ablation_loss}, we observe that our model outperforms the decoupled variant by a large margin for attribute prediction. This confirms the hypothesis that joint training of attribute prediction and link prediction performs better than training separately. We also see that sharing history for attribute prediction and link prediction deteriorates the results, which indicates that the history encoding information required for link prediction and attribute prediction is quite different from each other. Lastly, the time-independent variant of our framework performs poorly. This clearly indicates that the temporal evolution information is essential for proper inference.
\paragraph{Sensitivity analysis of hyperparameter $\lambda$.}
We perform the sensitivity analysis of parameter $\lambda$, which specifies the weight given to both the tasks. We show the variation of MSE loss for attribute prediction task with $\lambda$.
Figures \ref{lambad_variation_Loss} shows the variation of Attribute Loss with increasing $\lambda$. In Figure \ref{lambad_variation_Loss}, we observe that Attribute value decreases with increasing $\lambda$. As expected, as increasing lambda favors the optimization of attribute loss while decreasing lambda favors the link prediction.
\section{Conclusion and Future Work}
In this paper, we propose to jointly model the attribute prediction and link prediction on a temporally evolving graph. We propose a novel framework \frameworkName, which uses two recurrent neural networks to encode the history for the graph. The framework shares the parameter for the two tasks and jointly trains the two tasks using multi-task learning. Through various experiments, we show that our framework is able to achieve better performance on attribute prediction than the previous methods indicating that external knowledge is useful for time series prediction. Interesting future work includes the link prediction on graph level rather than on subject and relation level in a memory-optimized way.

\section*{Acknowledgements}
This research is based upon work supported in part by NSF SMA 18-29268 and United States Office Of Naval Research under Contract No. N660011924033.The views and conclusions contained herein are those of the authors and should not be interpreted as necessarily representing the official policies, either expressed or implied, of the U.S. Government. We would like to thank all the collaborators in USC INK research lab for their constructive feedback on the work.

\bibliography{08refs}

\begin{thebibliography}{}

\bibitem[\protect\citeauthoryear{Bordes \bgroup \em et al.\egroup
  }{2013}]{transE}
Antoine Bordes, Nicolas Usunier, Alberto Garcia-Duran, Jason Weston, and Oksana
  Yakhnenko.
\newblock Translating embeddings for modeling multi-relational data.
\newblock In C.~J.~C. Burges, L.~Bottou, M.~Welling, Z.~Ghahramani, and K.~Q.
  Weinberger, editors, {\em Advances in Neural Information Processing Systems
  26}, pages 2787--2795. Curran Associates, Inc., 2013.

\bibitem[\protect\citeauthoryear{Caruana}{1997}]{MTLRich}
Rich Caruana.
\newblock Multitask learning.
\newblock {\em Machine Learning}, 28(1):41--75, 1997.

\bibitem[\protect\citeauthoryear{Cho \bgroup \em et al.\egroup }{2014}]{GRU}
KyungHyun Cho, Bart van Merrienboer, Dzmitry Bahdanau, and Yoshua Bengio.
\newblock On the properties of neural machine translation: Encoder-decoder
  approaches.
\newblock {\em CoRR}, abs/1409.1259, 2014.

\bibitem[\protect\citeauthoryear{Dasgupta \bgroup \em et al.\egroup
  }{2018}]{hyte}
Shib~Sankar Dasgupta, Swayambhu~Nath Ray, and Partha Talukdar.
\newblock {H}y{TE}: Hyperplane-based temporally aware knowledge graph
  embedding.
\newblock In {\em Proceedings of the 2018 Conference on Empirical Methods in
  Natural Language Processing}, pages 2001--2011, Brussels, Belgium,
  October-November 2018. Association for Computational Linguistics.

\bibitem[\protect\citeauthoryear{Dettmers \bgroup \em et al.\egroup
  }{2017}]{convE}
Tim Dettmers, Pasquale Minervini, Pontus Stenetorp, and Sebastian Riedel.
\newblock Convolutional 2d knowledge graph embeddings.
\newblock {\em CoRR}, abs/1707.01476, 2017.

\bibitem[\protect\citeauthoryear{Garc{\'{\i}}a{-}Dur{\'{a}}n \bgroup \em et
  al.\egroup }{2018}]{TA-distmult}
Alberto Garc{\'{\i}}a{-}Dur{\'{a}}n, Sebastijan Dumancic, and Mathias Niepert.
\newblock Learning sequence encoders for temporal knowledge graph completion.
\newblock {\em CoRR}, abs/1809.03202, 2018.

\bibitem[\protect\citeauthoryear{Goyal \bgroup \em et al.\egroup
  }{2018}]{dynamicApp3}
Palash Goyal, Nitin Kamra, Xinran He, and Yan Liu.
\newblock Dyngem: Deep embedding method for dynamic graphs.
\newblock {\em CoRR}, abs/1805.11273, 2018.

\bibitem[\protect\citeauthoryear{Hamilton \bgroup \em et al.\egroup
  }{2017}]{RL1}
William~L. Hamilton, Rex Ying, and Jure Leskovec.
\newblock Representation learning on graphs: Methods and applications.
\newblock {\em CoRR}, abs/1709.05584, 2017.

\bibitem[\protect\citeauthoryear{Hamilton}{1994}]{VAR}
{James Douglas} Hamilton.
\newblock {\em Time series analysis}.
\newblock Princeton Univ. Press, Princeton, NJ, 1994.

\bibitem[\protect\citeauthoryear{Jin \bgroup \em et al.\egroup }{2019}]{renet}
Woojeong Jin, He~Jiang, Changlin Zhang, Pedro Szekely, and Xiang Ren.
\newblock Recurrent event network for reasoning over temporal knowledge graphs.
\newblock {\em arXiv preprint arXiv:1904.05530}, 2019.

\bibitem[\protect\citeauthoryear{Kipf and Welling}{2016}]{gcn}
Thomas~N. Kipf and Max Welling.
\newblock Semi-supervised classification with graph convolutional networks.
\newblock {\em CoRR}, abs/1609.02907, 2016.

\bibitem[\protect\citeauthoryear{Kipf \bgroup \em et al.\egroup }{2018}]{kg2}
Thomas~N. Kipf, Ethan Fetaya, Kuan-Chieh Wang, Max Welling, and Richard~S.
  Zemel.
\newblock Neural relational inference for interacting systems.
\newblock In {\em ICML}, 2018.

\bibitem[\protect\citeauthoryear{Kumar \bgroup \em et al.\egroup
  }{2018}]{dynamicApp2}
Srijan Kumar, Xikun Zhang, and Jure Leskovec.
\newblock Learning dynamic embeddings from temporal interactions.
\newblock {\em CoRR}, abs/1812.02289, 2018.

\bibitem[\protect\citeauthoryear{Li \bgroup \em et al.\egroup
  }{2017}]{DCRNN-YanLiu}
Yaguang Li, Rose Yu, Cyrus Shahabi, and Yan Liu.
\newblock Graph convolutional recurrent neural network: Data-driven traffic
  forecasting.
\newblock {\em CoRR}, abs/1707.01926, 2017.

\bibitem[\protect\citeauthoryear{Lippi \bgroup \em et al.\egroup
  }{2013}]{Lippi2013ShortTermTF}
Marco Lippi, Matteo Bertini, and Paolo Frasconi.
\newblock Short-term traffic flow forecasting: An experimental comparison of
  time-series analysis and supervised learning.
\newblock {\em IEEE Transactions on Intelligent Transportation Systems},
  14:871--882, 2013.

\bibitem[\protect\citeauthoryear{Liu \bgroup \em et al.\egroup
  }{2011}]{Liu:2011:DSC:2020408.2020571}
Wei Liu, Yu~Zheng, Sanjay Chawla, Jing Yuan, and Xie Xing.
\newblock Discovering spatio-temporal causal interactions in traffic data
  streams.
\newblock In {\em Proceedings of the 17th ACM SIGKDD International Conference
  on Knowledge Discovery and Data Mining}, KDD '11, pages 1010--1018, New York,
  NY, USA, 2011. ACM.

\bibitem[\protect\citeauthoryear{Neil \bgroup \em et al.\egroup }{2018}]{kg1}
Daniel Neil, Joss Briody, Alix Lacoste, Aaron Sim, Paidi Creed, and Amir
  Saffari.
\newblock Interpretable graph convolutional neural networks for inference on
  noisy knowledge graphs.
\newblock {\em CoRR}, abs/1812.00279, 2018.

\bibitem[\protect\citeauthoryear{Pareja \bgroup \em et al.\egroup
  }{2019}]{dynamicApp1}
Aldo Pareja, Giacomo Domeniconi, Jie Chen, Tengfei Ma, Toyotaro Suzumura,
  Hiroki Kanezashi, Tim Kaler, and Charles~E. Leisersen.
\newblock Evolvegcn: Evolving graph convolutional networks for dynamic graphs.
\newblock {\em CoRR}, abs/1902.10191, 2019.

\bibitem[\protect\citeauthoryear{Rossi}{2018}]{rossi_2018}
Ryan~A. Rossi.
\newblock Relational time series forecasting.
\newblock {\em The Knowledge Engineering Review}, 33:e1, 2018.

\bibitem[\protect\citeauthoryear{Ruder}{2017}]{MTLSurvey}
Sebastian Ruder.
\newblock An overview of multi-task learning in deep neural networks.
\newblock {\em CoRR}, abs/1706.05098, 2017.

\bibitem[\protect\citeauthoryear{Schlichtkrull \bgroup \em et al.\egroup
  }{2018}]{RGCN}
Michael Schlichtkrull, Thomas~N. Kipf, Peter Bloem, Rianne van~den Berg, Ivan
  Titov, and Max Welling.
\newblock Modeling relational data with graph convolutional networks.
\newblock {\em Lecture Notes in Computer Science}, page 593–607, 2018.

\bibitem[\protect\citeauthoryear{Sutskever \bgroup \em et al.\egroup
  }{2014}]{seq2seq}
Ilya Sutskever, Oriol Vinyals, and Quoc~V. Le.
\newblock Sequence to sequence learning with neural networks.
\newblock {\em CoRR}, abs/1409.3215, 2014.

\bibitem[\protect\citeauthoryear{Trivedi \bgroup \em et al.\egroup
  }{2017}]{know-evolve}
Rakshit Trivedi, Hanjun Dai, Yichen Wang, and Le~Song.
\newblock Know-evolve: Deep reasoning in temporal knowledge graphs.
\newblock {\em CoRR}, abs/1705.05742, 2017.

\bibitem[\protect\citeauthoryear{Trivedi \bgroup \em et al.\egroup
  }{2018}]{dyrep}
Rakshit Trivedi, Mehrdad Farajtabar, Prasenjeet Biswal, and Hongyuan Zha.
\newblock Representation learning over dynamic graphs.
\newblock {\em CoRR}, abs/1803.04051, 2018.

\bibitem[\protect\citeauthoryear{Wu \bgroup \em et al.\egroup }{2019}]{survey1}
Zonghan Wu, Shirui Pan, Fengwen Chen, Guodong Long, Chengqi Zhang, and
  Philip~S. Yu.
\newblock A comprehensive survey on graph neural networks.
\newblock {\em CoRR}, abs/1901.00596, 2019.

\bibitem[\protect\citeauthoryear{Yang \bgroup \em et al.\egroup
  }{2014}]{DistMult}
Bishan Yang, Wen tau Yih, Xiaodong He, Jianfeng Gao, and Li~Deng.
\newblock Embedding entities and relations for learning and inference in
  knowledge bases.
\newblock {\em CoRR}, abs/1412.6575, 2014.

\bibitem[\protect\citeauthoryear{You \bgroup \em et al.\egroup
  }{2018}]{graphrnn}
Jiaxuan You, Rex Ying, Xiang Ren, William~L. Hamilton, and Jure Leskovec.
\newblock Graphrnn: {A} deep generative model for graphs.
\newblock {\em CoRR}, abs/1802.08773, 2018.

\end{thebibliography}
\bibliographystyle{named}

\clearpage
\appendix
\section*{Appendix}
\section{Datasets}
Due to the unavailability of datasets satisfying our problem statement we curated appropriate datasets by scraping the web. We created and tested our approach on the datasets described below.

\para{Attributed Trade graph (ATG).} This dataset consists of a directed, multi-relational, unweighted, dynamic knowledge graph with nodes representing different countries. A timestamped edge between two nodes represents the net exports between the respective countries in million dollars. To discretize the edges, the range of values of net exports is split into 200 equal-sized segments resulting in 178 different types of edges. The attribute value associated with each node is the month-averaged currency exchange rate of the corresponding country in SDRs per currency unit. The data is present in the form of a tuple $(h,r,t,a_h,a_t,\tau)$ where $h,t,\tau$ denote the head, tail and timestamp respectively. Relation $r$ exists between $h$ and $t$ at timestamp $\tau$ and $a_h, a_t$ are the attribute values of the head and tail respectively at $\tau$. The graph evolves at a monthly rate.\\
The graph is obtained by using a script to scrape data from www.trademap.org. The exchange-rate data is scraped from www.imf.org.

\para{Co-authorship-Citation dataset (CAC).} Here the knowledge graph is dynamic, uni-relational, unweighted and undirected. The nodes denote authors and an edge between two nodes at a particular timestamp denotes that the corresponding authors contributed to a research paper at that time. The time granularity is a year. The attribute value associated with each node is the number of citations received by the associated author on any paper written by him/her per year. Again the data is present in the form of the tuple $(h,r,t,a_h,a_t,\tau)$ where the meanings of the symbols are as explained above. We used two versions of this dataset: small having 44 nodes and large having 20k nodes.\\
The citation dataset is curated from www.aminer.cn.

\para{Multi-attributed Trade graph (MTG).} The graph in this dataset is a subset of the trade graph described above. Each node has multiple attribute values associated with it. One of the two attributes is the Net Export Price Index with individual commodities weighted by the ratio of net exports to the total commodity trade. The other is the value of International Reserves and other foreign currency assets in millions of US dollars. All two form monthly time series. \\
Both the time series attributes are scraped from www.imf.org.

\para{Attributed GDELT graph (AGG).} The knowledge graph, in this case, is derived from the Global Database of Events, Language, and Tone (GDELT). It is dynamic, directed, multi-relational, unweighted and also has multiple types of nodes. The nodes represent entities like political leaders, organisations and several others. Each of these entities can be associated with a country. We modified this graph by adding nodes representing countries and connecting them to their respective entities through a self-defined edge type. 245 other edge types also exist recording events. We have used this graph at the granularity level of a month. Only the country nodes are associated with a time-series attribute which is taken as the Currency Exchange Rate (as described above) in this case.

\section{Experimental Settings}
All are models are written in PyTorch\footnote{https://pytorch.org/}. We use the Adam Optimizer for training our models with learning rate of $10^{-3}$. The Gated recurrent units are used as the RNN for all the experiments. We use only one GRU unit with $200$ hidden dimension for the experiments involving knowledge graphs while we use both one unit and two units for the baselines. The default sequence length for input to the graph is used. We experiment with various values of $\lambda$ and we report the results of study in Section \ref{Ablation Description}. We first train \frameworkName on the training dataset. We then use the saved checkpoints from various stages of the training to obtain the results of attribute prediction on the validation data. From the validation results, we choose the best checkpoint and evaluate the test set on that checkpoint. We report the results of attribute prediction on the test data. All are models are trained on Nvidia GeForce GTX 1080 Ti.

\end{document}